\documentclass[utf8]{FrontiersinHarvard} 

\usepackage{url,hyperref,lineno,microtype,subcaption}
\usepackage{graphicx}
\usepackage{amsmath}
\usepackage{amssymb}
\usepackage{booktabs}
\usepackage{caption}
\usepackage{subcaption}
\usepackage{multicol}
\usepackage{array}
\usepackage[onehalfspacing]{setspace}

\def\keyFont{\fontsize{8}{11}\helveticabold }
\def\firstAuthorLast{Johnson {et~al.}} 
\def\Authors{Faith Johnson\,$^{1,*}$, Jack Lowry\,$^{1}$, Kristin Dana\,$^{1}$ and Peter Oudemans\,$^{2,3}$}
\def\Address{$^{1}$ ECE Department, Rutgers University, New Brunswick, NJ, USA \\
$^{2}$ Department of Plant Biology, Rutgers University, New Brunswick, NJ, USA \\
$^{3}$ Philip E. Marucci Center for Blueberry and Cranberry Research and Extension, Rutgers University, Chatsworth, NJ, USA}
\def\corrAuthor{Faith Johnson}

\def\corrEmail{faith.johnson@rutgers.edu}

\begin{document}
\onecolumn
\firstpage{1}

\title {Vision-Based Cranberry Crop Ripening Assessment} 

\author[\firstAuthorLast ]{\Authors} 
\address{} 
\correspondance{} 

\extraAuth{}

\maketitle

\begin{abstract}

Agricultural domains are being transformed by recent advances in  AI and computer vision that support quantitative visual evaluation. Using drone imaging, we develop a framework for characterizing the ripening process of cranberry crops. Our method consists of drone-based time-series collection over a cranberry growing season, photometric calibration for albedo recovery from pixels, and berry segmentation with semi-supervised deep learning networks using point-click annotations. By extracting time-series berry albedo measurements, we evaluate four different varieties of cranberries and provide a quantification of their ripening rates.  Such quantification has practical implications for 1) assessing real-time overheating risks for cranberry bogs; 2) large scale comparisons of progeny in crop breeding; 3) detecting disease by looking for ripening pattern outliers.  This work is the first of its kind in quantitative evaluation of ripening using computer vision methods and has impact beyond cranberry crops including wine grapes, olives, blueberries, and maize.  

\tiny
 \keyFont{ \section{Keywords:} high throughput phenotyping, albedo analysis, semantic segmentation, crop yield estimation, counting methods} 
 
\end{abstract}




\section{Introduction}

Machine learning and computer vision methods  play an increasingly vital role in facilitating agricultural advancement by giving real time, actionable crop feedback \cite{luo2023semantic,yin2022computer,meshram2021machine}.  
These methods are enabling farming practices  to adapt and evolve to keep up with changing conditions. 
Cranberry farmers are particularly poised to benefit from vision-based crop monitoring 
as they face
numerous challenges related to fruit quality such as fruit rot and over heating \cite{oudemans1998cranberry, polashock2009north, vorsa2012american, vorsa2019domestication}. 
As cranberries ripen and turn red, they become much more susceptible to overheating, partially because they lose their capacity for evaporative cooling \cite{kerry2017investigating,racsko2012sunburn,smart1976solar}. When this growth stage is reached, 
the cranberries exposed to direct sunlight can overheat and become unusable.

\begin{figure}
    \centering
    \includegraphics[width=0.23\textwidth]{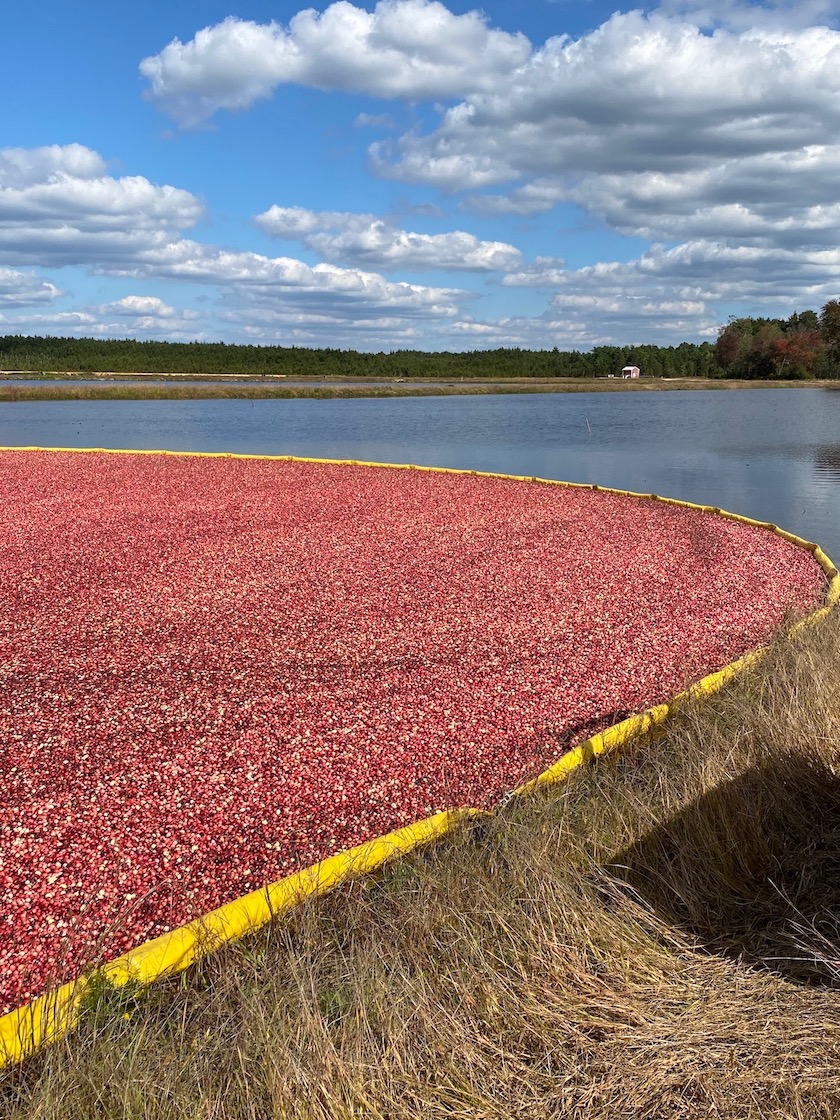}
    \includegraphics[width=0.225\textwidth]{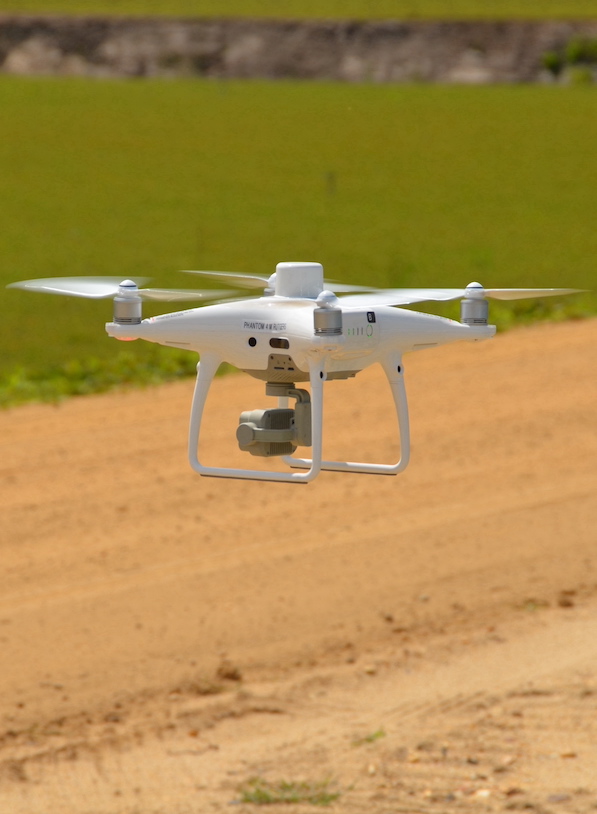}
    \caption{Cranberry bog at the measurement site. (Left) Cranberry harvesting. (Right) Drone at bog for in-field cranberry measurements during the growing season. 
    }
    \label{fig:teaser}
\end{figure}

We develop a  vision-based method for measuring in-field cranberry albedo to quantify ripening in order to predict when cranberries are nearing this vulnerable stage. 
Currently, cranberry growers quantify this ripening process by using out-of-field  albedo evaluation by imaging harvested cranberries over time \cite{oceanSprayPerCom}.
This approach is cumbersome and time-consuming, limiting its utility in larger-scale evaluations. For practical application, only small numbers of berries can be harvested for out-of-field images.  
Furthermore, since berries can overheat in a short period of time \cite{pelletier2016reducing,kerry2017investigating}, irrigation decisions should be coordinated with in-field albedo characterization, informing the grower on the number of vulnerable berries and enabling expedient decision making.  
%

The conventional solution to overheating is increased crop irrigation during the growing season. 
Inadequate or poorly timed irrigation can lead to overheating whereas excessive irrigation encourages fungal fruit rot to develop
\cite{oudemans1998cranberry}. 
Irrigation decisions must also consider cost and efficient use of environmental resources. 
Berries on the top of the canopy with direct sun exposure have a high risk of overheating, while berries underneath the leafy canopy are generally well protected. Therefore, assessing the current albedo of the visible berries is directly relevant to irrigation decisions.
Ripening rates vary among cranberry varieties, and ones that ripen early are at the greatest risk.
In-field measurement of the ripening rate for a particular cranberry bed
significantly informs crop management decisions. 


Our ripening assessment framework uses cranberry image segmentation to evaluate albedo variation over time to compare cranberry varieties.
In recent prior work \cite{akiva2022vision},  neural networks for segmentation have been used for yield estimation through counting. In our work, we use these segmentation networks to isolate individual berries over time to find temporal patterns in cranberry albedo across varieties. 
By combining counting and albedo analysis, 
it becomes possible to evaluate the economic risk to a particular crop on a high temperature day, while also making a long term assessment on which varieties provide the most yield.

The cranberries are segmented using a  semi-supervised method \cite{akiva2022vision} based on the Triple-S network \cite{akiva2020finding}. Using point-wise annotations instead of pixel-wise labels significantly reduces the labeling cost for the densely populated cranberry images. 
We provide new point-click labeled cranberry imagery that supports ripening assessments in CRAID-2022, a new dataset covering a larger time period of the growing season with more temporal frequency than prior work.
We train a segmentation network to isolate cranberry pixels from drone imagery in both this new dataset and the preexisting CRAID-2019 dataset \cite{akiva2020finding}. 
We present a  ripening comparison of four cranberry varieties over a two month span and show clear  timelines of albedo change indicating when each variety becomes at great risk for overheating.

\begin{figure}
    \centering
   \includegraphics[width=0.5\textwidth]{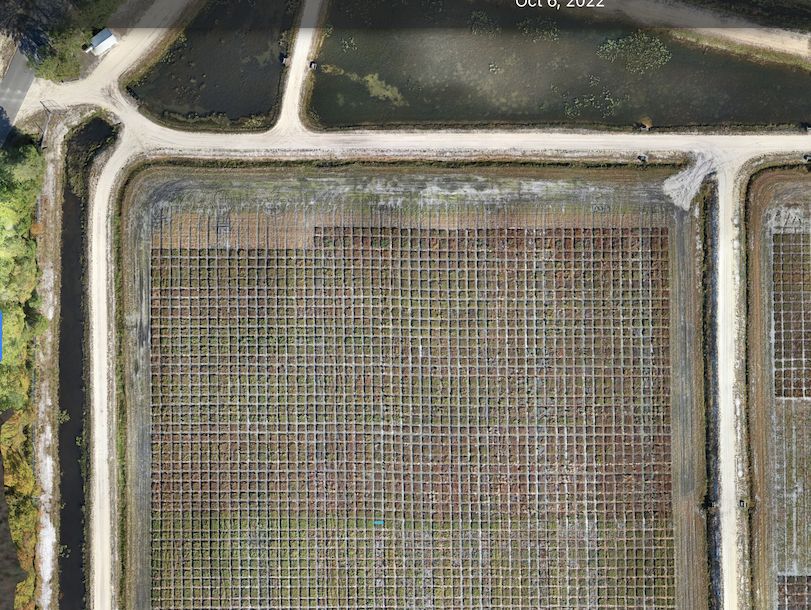}
    \caption{An example of breeding plots (drone view) that are typically evaluated manually.  Planting design permits approx 3500 plots/ha, and this entire block is approximately 2 ha.  Convenient quantitative evaluation can be supported by our vision-based ripening assessment framework. {\it Location removed for blind review.}}
    \label{fig:HTP}
\end{figure}

\subsection{Impact for Crop Breeding} Screening for the heritability of novel genotypes requires high through-put phenotyping (HTP) methods to discover desirable genetic traits \citep{araus2014field, diaz2018massive}. In crop breeding, there may be hundreds to thousands of progeny/offspring to evaluate, and high throughput methods make this evaluation practical. Computer vision algorithms for segmentation and calibrated albedo measurements enable quantitative comparisons.  The  methodology we present in this paper fits those requirements well, and HTP is an application domain for this work.  

The rate of color development is a crop trait that can affect the quality of cranberries at harvest. For consumer appeal,  the timing and uniformity of ripening is critical, i.e. asynchronous ripening is a problem. For breeding, uniformity is desirable so   HTP is used to look at multiple genotypes. For example, our related current work (unpublished) evaluates 300-400 genotypes planted in small plots (e.g. 3.3 sq.\textbackslash m.) where a ripening evaluation is done out-of-field and only a few times (1-2) per season depending on time and labor.  To illustrate the scale of these studies, consider that they include 0.5 acre plots  with 350 individual small plots and 
0.2 hectare (ha) plots with 7000 individual plots (see drone image shown in Figure~\ref{fig:HTP}).

\section{Related Work}



\subsection{Precision Agriculture}

Precision agriculture is revolutionizing farming and challenging traditional methods. Future farms will integrate multiple technical advances such as soil sensors \cite{abdollahi2021wireless}, plant wearables \cite{yin2021soil}, drone aerodynamics \cite{radoglou2020compilation}, and remote sensing \cite{sishodia2020applications}.
Advances in machine learning and AI have been particularly impactful, enabling significant breakthroughs in agricultural applications in recent years \citep{wang2022review,benos2021machine,sharma2020machine,mavridou2019machine}. Computer vision is capable of giving real time, high fidelity feedback to farmers about yield estimation \citep{van2020crop,darwin2021recognition,he2022fruit,palacios2023early}, phenotype identification \citep{li2020review,kolhar2023plant,liu2022tomatodet}, and crop health assessment \citep{dhaka2021survey,ahmad2022survey,kattenborn2021review} while also being useful for larger scale applications like farm automation \cite{friha2021internet}.


\subsection{Weakly Supervised Semantic Segmentation}
Semantic segmentation is an extremely useful tool for applications that require object localization or counting \citep{jia2022accurate,ilyas2021dam,afonso2020tomato}, like crop yield estimation \cite{maheswari2021intelligent}. However, obtaining enough pixel-wise labels to effectively train a network on images with high label densities, as is the case in the cranberry domain, is prohibitively expensive. To combat this, some work uses image level labels \citep{araslanov2020single,cermelli2022incremental} to guide the segmentation, but performance significantly improves with instance-level information \cite{zhou2018weakly}. 

A middle ground between these two approaches is point-wise supervision \citep{cheng2022pointly,liu2021one,song2021rethinking}, which provides point-level localization for each class instance. 
For the specific domain of cranberry cultivation, Akiva et al. \citep{akiva2022vision,akiva2021ai,akiva2020finding} create a weakly supervised pipeline for semantic segmentation and counting of cranberries from aerial drone images. This method utilizes point-wise labels to supervise their network, negating the need for costly pixel-wise annotations in densely populated scenes. While their work focuses on yield estimation and crop risk assessment, our work utilizes a modification of their pipeline for  albedo characterization in order to analyze and compare cranberry albedo over time and across different species.

\subsection{Albedo Characterization over Time}

 As cranberries ripen, their risk of spoilage increases due to overheating \cite{pelletier2016reducing} caused by a decrease in evapotranspiration. 
  This ripening corresponds with visual changes in the berry albedo, which allows us to use albedo characterization over time to predict when a cranberry bog is most at risk.
 This same phenomena is also found in apples \cite{racsko2012sunburn} and grapes \cite{smart1976solar}.
 Ripening patterns can also indicate the presence of viruses as occurs in wine grapes \citep{alabi2016impacts,blanco2017red}.
  Despite the importance of quantifying color development, automated methods for albedo characterization have received limited attention in the literature. 
 Most existing studies of ripening, do out-of-field measurements that rely on harvested berries for evaluating ripening \citep{vorsa2017performance, keller2010managing}. 
 These methods  are time-consuming and do not scale to large evaluations or real-time assessments. 
  The framework of this paper is an important step for using advanced computer vision algorithms (semi-supervised segmentation with low-cost annotation, deep learning networks) as a tool in agriculture. 



\begin{figure} 

    \includegraphics[width=\textwidth]{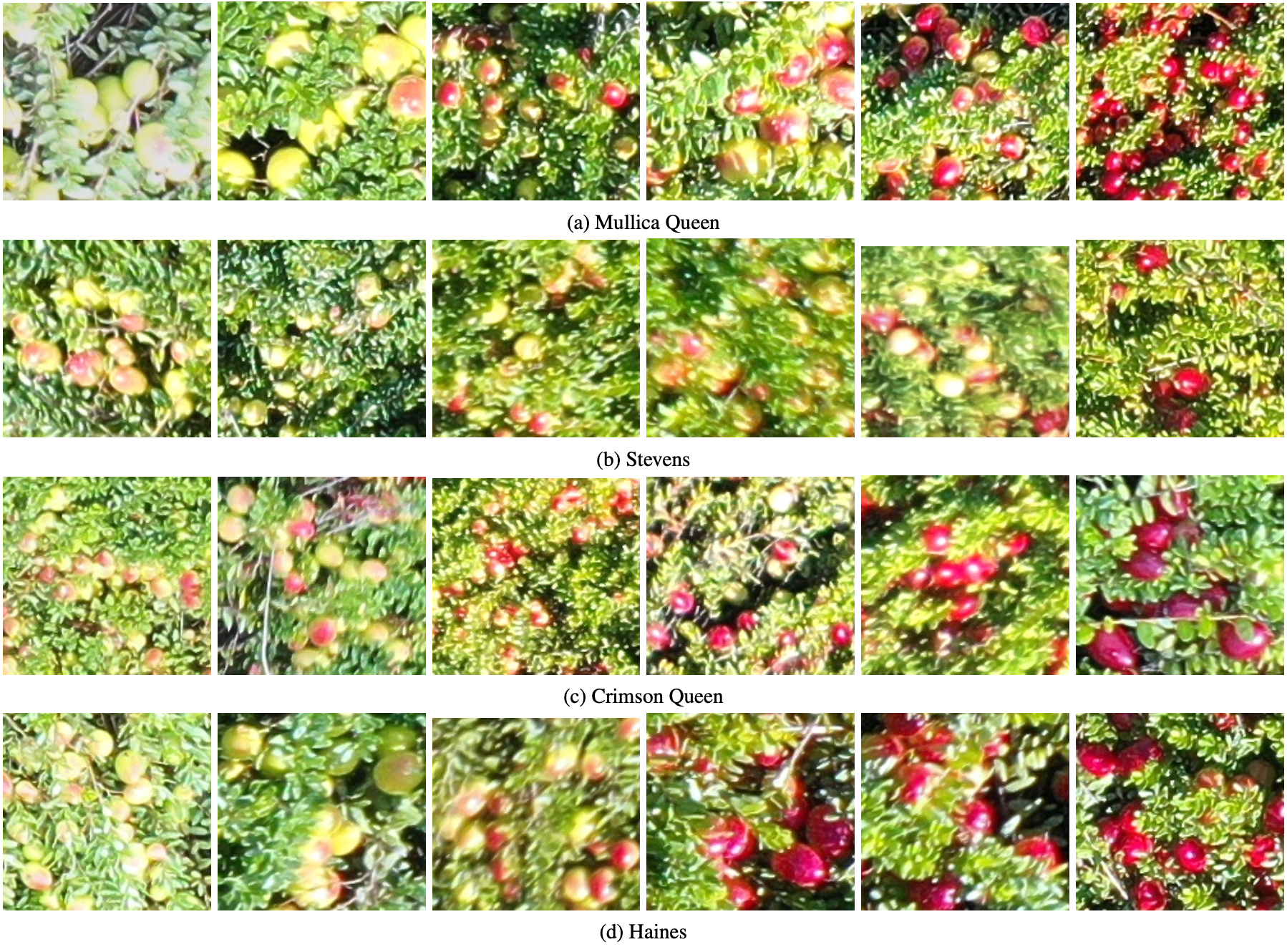}
    
    \caption{Drone images from multi-temporal drone-scout imaging. Weekly inspection of multiple cranberry bogs over the late July/September growing season for four varieties (Mullica Queen, Stevens, Crimson Queen, Haines). (Left to Right) Imaging Dates for 2022: 7/27, 8/2, 8/16, 8/25, 8/31, 9/9.  }

    \label{fig:my_label}
\end{figure}

\begin{table}[]
    \centering
    \renewcommand{\arraystretch}{1.5}
    \begin{tabular}{| c c| c c c | c c |}
    \hline 
      Dataset & & & CRAID-2022 & & & CRAID-2019 \\
    \hline 
      Total Number of Images & & & 7,198 & & & 21,436 \\
      Number of Labelled Images & & & 220 & & & 2368 \\
      Date Range of Collection & & & 07/27/22 - 09/09/22 & & & 08/20/19 - 10/20/19 \\
      Collection Frequency (times per week) & & & 1 & & & 3 \\
      Sensors & & & RGB & & & RGB \\
      Image Size $(\text{px}^2)$ & & & 456 x 608 & & & 456 x 608\\
    \hline
    \end{tabular}
    \caption{Details of the CRAID-2022 dataset. Images were collected via drone with an RGB sensor of multiple cranberry bogs over the late July - September growing season. In total, there were four varieties (Mullica Queen, Stevens, Crimson Queen, Haines) imaged over six dates roughly a week apart. We compare collection details between CRAID-2022 and CRAID 2019 for convenience. }
    \label{tab:my_label}
\end{table}

\section{Datasets}

For data collection, we introduce CRAID-2022, obtained from our bog monitoring system at PE Marucci Center for Blueberry and Cranberry Research, a substation of the Rutgers New Jersey Agricultural
Experiment Station (Chatsworth, NJ). To the best of our knowledge,
this is the first dataset and methodology that support end-to-end berry
health assessment from sky and ground data. This framework is a
pioneering step in the area of vision/AI for precision agriculture and
especially a precision-based method for rapid, short-term decision making that can assist growers to implement irrigation practices in response
to complex risk factors.
We combine the  CRAID-2019 dataset \cite{akiva2020finding} with drone-based cranberry bog images from the 2022 growing season. In this paper, we will use the term CRAID$+$ dataset to refer to the combination of CRAID 2019 and CRAID 2022.
In total, this dataset contains four different cranberry varieties over seven bogs. 
The images include three beds of Mullica Queen, one bed of Stevens, two beds of Haines, and one bed of Crimson Queen cranberries. The new images were taken by drone in weekly increments between the months of July and September. 

We calibrate each drone image and crop each into 72, non-overlapping $456 \times 608$ sub-images used as training images. A selection of 220 crops representative of the diverse berry appearances in the entire growing season were manually labelled with point-wise annotations for all berries in the image. This data was combined with the labeled dataset of 2368 images from \cite{akiva2020finding} to create a new training dataset comprised of 2588 total images.
We train on this combined dataset of 2588 images (CRAID$+$) as well as the smaller dataset of 220 images (CRAID-2022 only)
and compare the pipeline performance between the two.



\section{Methods}

\begin{figure*}
    \centering
    \includegraphics[width=0.33\textwidth]{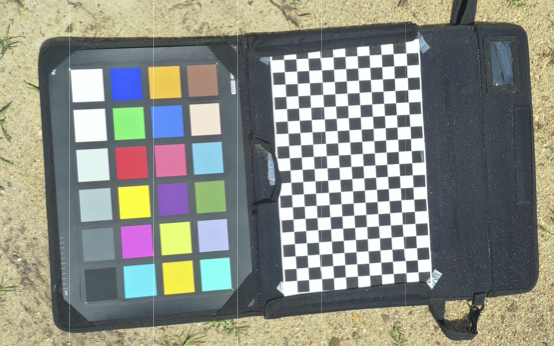}
    \includegraphics[width=0.31\textwidth]{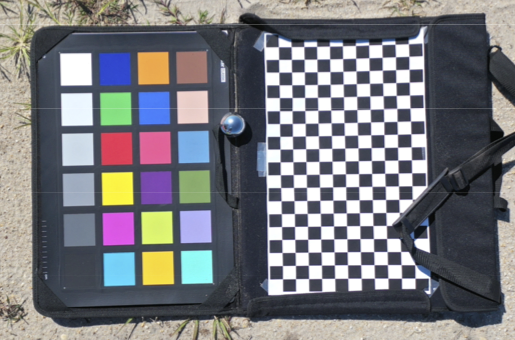}
    \includegraphics[width=0.33\textwidth]{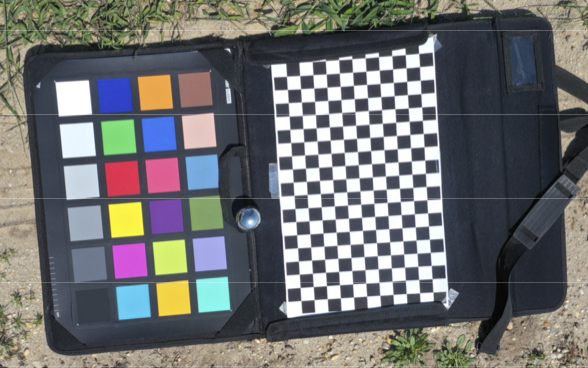}
    \caption{Images for photometric calibration. The drone imaging protocol includes images of  the Macbeth card over multiple days. Although the same camera is used for imaging, camera parameters change.
    Notice slight variations between the cards from different days, which we remove through photometric calibration.}
    \label{fig:macbeth}
\end{figure*}

\subsection{Photometric Calibration}
The images in CRAID-2022 from the 2022 growing season were first photometrically calibrated using the Macbeth Color Checker card and a well-established approach of estimating the optimal radiometric correction using measurements of the card under uniform illumination \citep{kim2008robust, debevec2008recovering,mitsunaga1999radiometric}. 
Radiometric or photometric calibration is needed to calibrate for the effects of the changing camera parameters between imaging sessions. Additionally, the change in sun angle will affect the appearance. Since our goal is to measure an invariant albedo measurement, raw pixels are insufficient since they depend on camera parameters and environment conditions. 
Reference images of the card were taken from every bog for each day of data collection using the drone camera. For each reference image we extracted intensity values for the 6 grey scale squares on the Macbeth Color Checker. The measured values were used to find a linear transformation to recover the radiometric correction parameters, and the images were corrected accordingly. 


\subsection{Segmentation Methods}
We use weakly-supervised semantic segmentation \cite{akiva2022vision} to isolate/segment cranberries in aerial images.  An image $X$ is fed into an autoencoder and the  output of the decoder is fed into three different branches. The first branch computes segmentation loss on the decoder output using point-wise annotations and a pseudo-mask generated from the features to push the network to localize cranberry instances. The second computes convexity loss on the predicted cranberry instances to make the predicted blobs round in shape. The final branch computes split loss to push the network toward predicting separable cranberry instances. Training semantic segmentation for this task using the traditional pixel-wise labels would require extensive labeling of images with high label densities. Using point-wise annotations (point-clicks instead of full segmentation ground truth) significantly lessens the labeling cost while still allowing for the accurate localization  of the cranberries.

Our investigation of this segmentation architecture for the task at hand led to several questions: Would the method work without training on new data since the cranberry crops were similar in previous and new datasets? Or, could the same architecture and losses be used with a new training set? Our empirical investigation led to the result that the same architecure could be re-used but new training was necessary. However, for the task of identifying cranberry positions, significantly less training data (10x less) was needed (See Section~\ref{sec:results}).

\subsection{Albedo}
The industry standard for ripeness defines five classes of cranberries based on albedo \cite{oceanSprayPerCom}. 
These distinct stages of ripeness are hand defined by field experts from the periodic collections of the cranberries throughout the season \cite{oceanSprayPerCom}. We use a similar approach for classifying albedo change in the CRAID data, but opt for using images of the cranberries in the bog instead. Each class is defined by TSNE clustering the RGB pixel values of a collection of randomly sampled cranberry detections from the entirety of the images from the 2022 growing season. We cluster points in this embedding space with k-means, then map those clusters to the 5 ``common classes", spanning from green to red, that best align with the industry standard. 

Once the classes have been determined, we match each berry to its corresponding color class by matching each pixel belonging to a single berry to its closest color cluster. The cluster belonging to the majority of the pixel values is chosen as the label for the berry. We repeat this process for each cranberry and count the number of detections in each image. The change in class density is plotted, as in Figure \ref{fig:albedoOverTime}, and clearly shows patterns in the cranberry albedo. From the progression of these plots, we also pinpoint when each cranberry variety becomes most at risk of overheating. 


\begin{figure}
    \centering
    \includegraphics[width=0.85\textwidth]{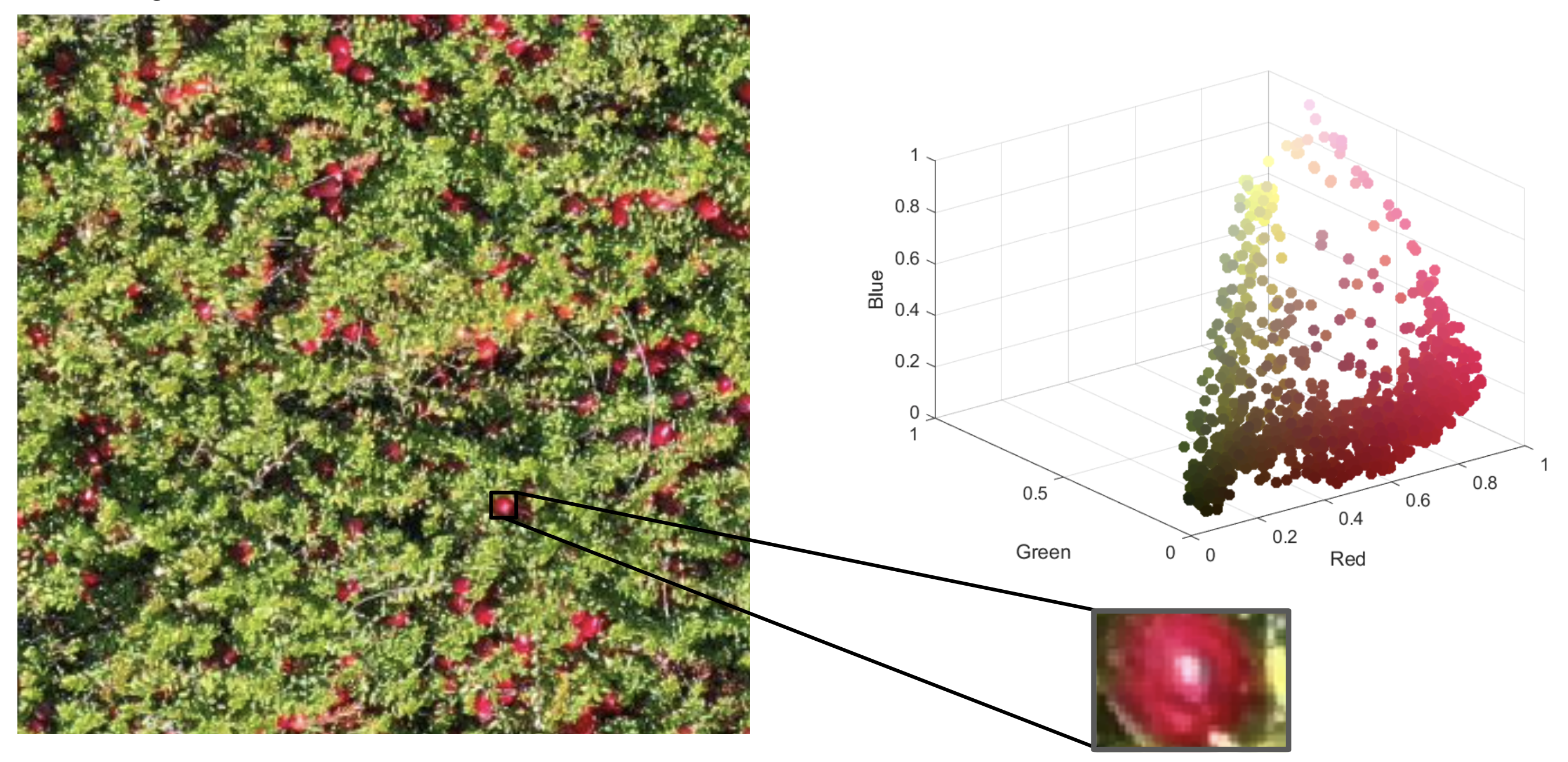}
    \caption{3D RGB plots of the berry colors in a patch around a berry. The color plot illustrates the color variety of a single berry including lighter colors near the specularity. This plot depicts a patch around a cranberry and includes background leaf pixels. Segmentation of berries is used to isolate berry pixels, ideally excluding the contribution of leaf pixels. The observed  deep red color of the ripe cranberry indicates a high overheating risk.}
    \label{fig:rgbAlbedo}
\end{figure}

\section{Results}
\label{sec:results}

\begin{figure}
    \centering


    \includegraphics[width=\textwidth]{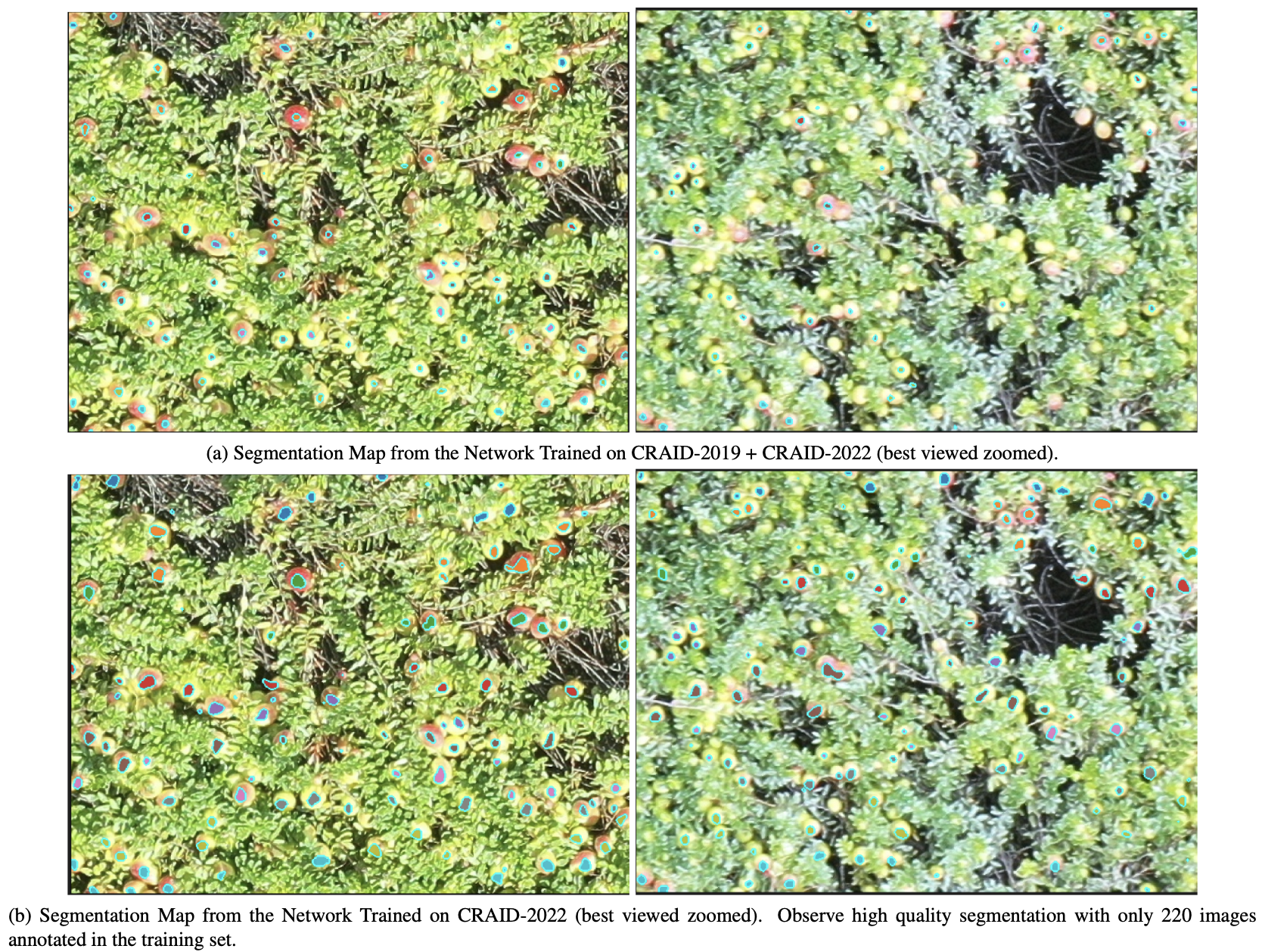}

    \caption{Example predicted cranberry segmentation maps overlaid on the original input images. (Best viewed zoomed.) The top row was predicted from the model trained on CRAID$+$. The bottom row was predicted by the model trained solely on the 2022 growing season data, CRAID-2022. The first model (first row) predicts significantly smaller cranberry blobs than the second model (second row). Additionally, the model trained on CRAID$+$ struggles to segment greener berries due to a lack of early images with green berries in the CRAID-2019 dataset that forms the majority of CRAID$+$ dataset. }
    \label{fig:segMap}
    
\end{figure}

\subsection{Cranberry Segmentation}

The cranberry segmentation network has a mean intersection over union (mIOU) of 62.54\% and a mean absolute error (MAE) of 13.46 as reported in \cite{akiva2021ai}. 
We show the results of the cranberry segmentation from the model trained on CRAID$+$ in Figure \ref{fig:segMap}a alongside the results from the model trained on only the 2022 growing season data (CRAID-2022) in Figure \ref{fig:segMap}b. Training on the larger CRAID$+$ dataset produces smaller predicted cranberry blobs than training on only the 2022 season data in CRAID-2022. This may be due to a scale mismatch between the datasets. The CRAID-2019 dataset contained images of cranberries taken by drone from a higher elevation than in the CRAID-2022 data. Another deficiency of the model trained on the CRAID$+$ data is that it misses detections of greener berries. Because the original CRAID-2019 dataset contains mostly red berries, it is unable to make the color invariant predictions necessary to accurately segment green berries in the CRAID-2022 dataset.
For this reason, we use the segmentation network trained on CRAID-2022 for our albedo characterization.

\begin{table}[]
    \centering
    \begin{tabular}{|c|c|c|c|c|c|c|}
    \hline
        \multicolumn{7}{|c|}{Ripeness Ratio} \\
        \hline 
        Bog  & 8/2 & 8/16 & 8/25 & 8/31 & 9/9 & 9/14 \\
        \hline
        \hline 
         A5 & 0.007 & 0.082 & 0.331 & 0.497 & 0.902 & 1 \\
         I15 & 0.001 & 0.108 & 0.167 & 0.409 & 0.874 & 1 \\
         J12 & 0.002 & 0.088 & 0.419 & 0.609 & 0.968 & 1\\
         K4 & 0.012 & 0.151 & 0.339 & 0.433 & 0.872 & 1 \\
         A4 & 0.127 & 0.453 & 0.926 & 1.118 & 0.808 & 1 \\
         B7 & 0.035 & 0.217 & 0.622 & 0.798 & 1.119 & 1 \\
         I3 & 0.010 & 0.079 & 0.347 & 0.678 & 1.121 & 1 \\
         \hline
    \end{tabular}
    \caption{Ripeness ratio for each bog of cranberries over time. We define the ripeness ratio for a bog of cranberries to be the percentage of red berries at the current time over the percentage of red berries on the final collection date. 
    Bog key indicates the following cranberry types: A5 Mullica Queen, I5 Mullica Queen, J12 Mullica Queen, K4 Stevens,  A4 Crimson Queen, B7 Haines, I3 Haines. 
    }
    \label{tab:ripenessRatio}
\end{table}

\subsection{Albedo Analysis}


Figure \ref{fig:rgbAlbedo} illustrates the process by which the cranberries are classified into one of the five common albedo classes. Once the berries are successfully segmented, their constituent pixels are matched to the closest color cluster. The berry is labeled with the cluster that appears most frequently. Figure \ref{fig:rgbAlbedo} shows the isolation of a cranberry and the distribution of the RGB values belonging to it. This berry is primarily red, so it would be mapped to class 5, which contains the reddest, most ripe berries.

\begin{figure}
    \centering

    \includegraphics[width=\textwidth]{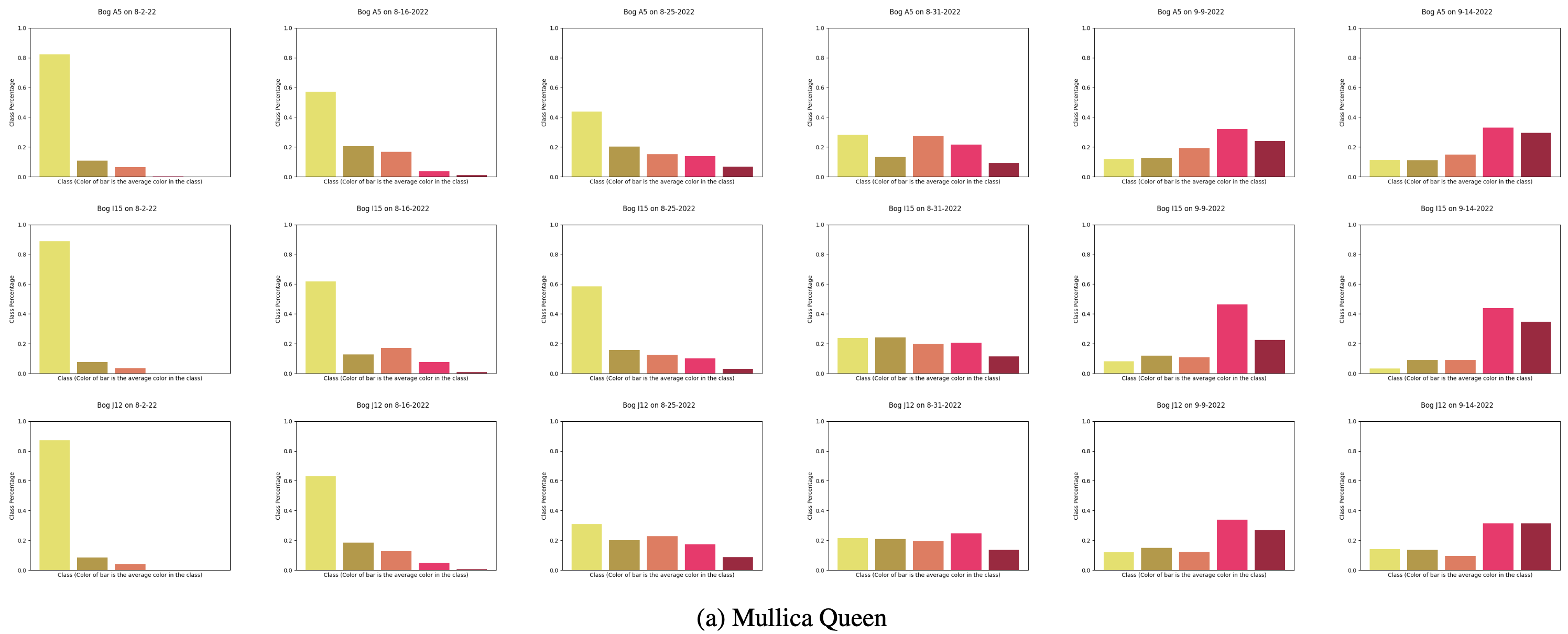}
    \includegraphics[width=\textwidth]{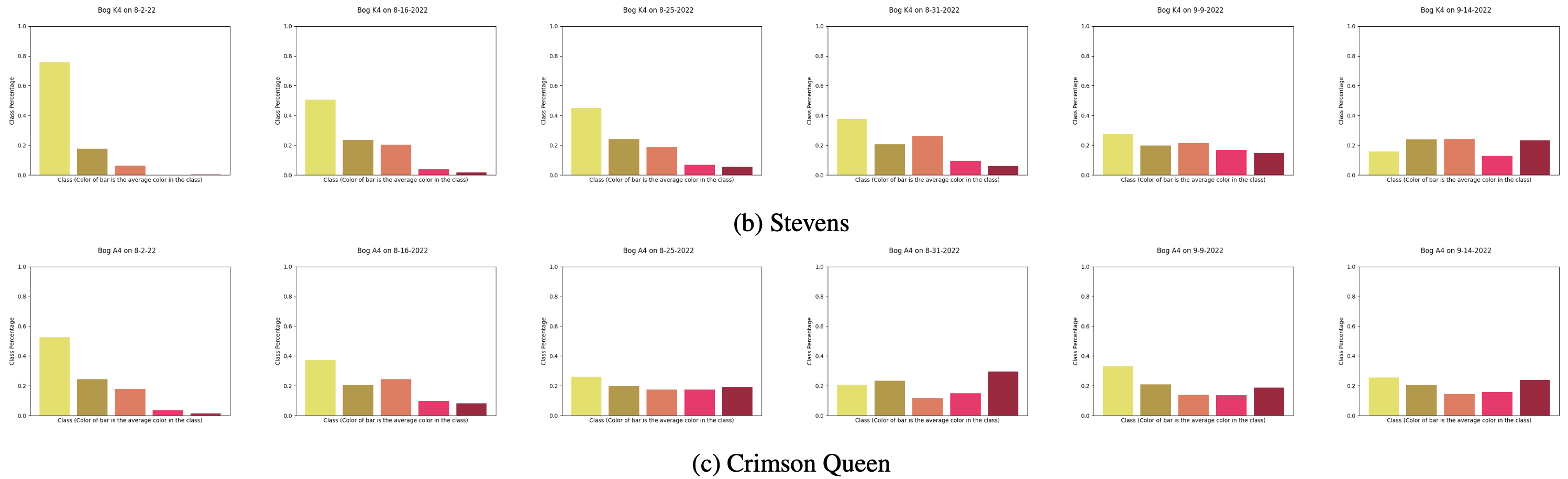}
    \includegraphics[width=\textwidth]{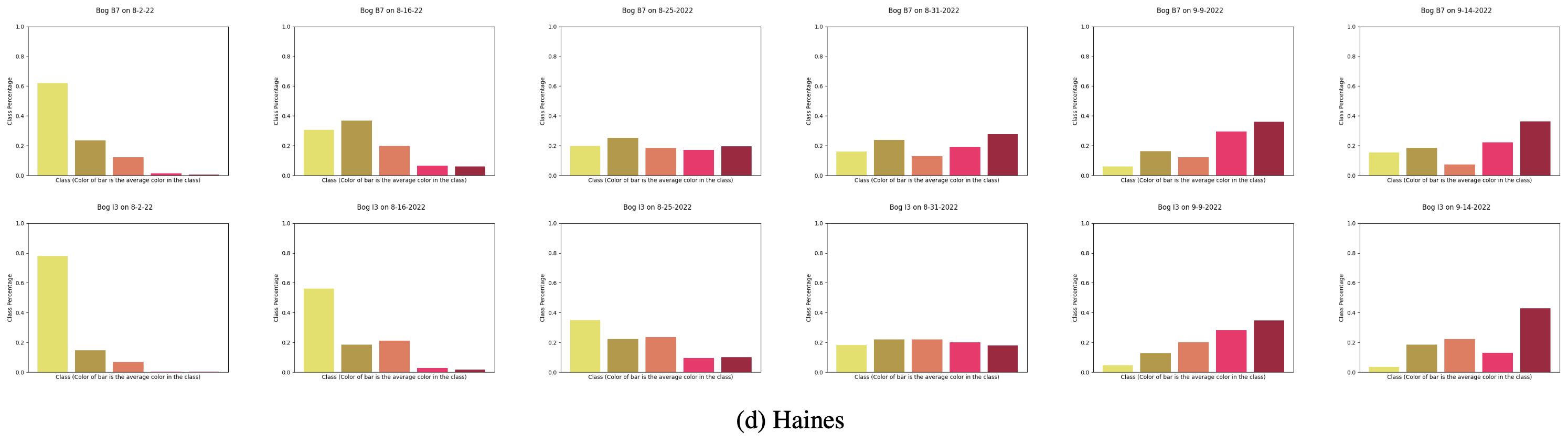}
    
    \caption{Plots comparing albedo over time for four cranberry varieties. Histograms of pixels in the five main color classes are shown. The four varieties are: Mullica Queen (top three rows), Stevens, Crimson Queen, and Haines (bottom two rows). Residual green pixels at the later dates are artifacts due to some misclassifications of background leaf pixels. }
    \label{fig:albedoOverTime}
\end{figure}

Once we compute the classes of all the berries, we plot the percentage of berries in each class for a particular collection date and bog in Figure \ref{fig:albedoOverTime}. The top three rows are the Mullica Queen variety. The next row is the Stevens variety followed by the Crimson Queen variety. The final two rows are the Haines variety. Each column of graphs is made from berries imaged on a specific day. From left to right, the columns were imaged on 8/2, 8/16, 8/25, 8/31, 9/9, and 9/14 in 2022.  As the berries redden, 
 the cranberry bog enters the high risk category and the 
the ripeness ratio, as shown in Table \ref{tab:ripenessRatio}, can be used to determine a
ripeness threshold (e.g.  approximately 0.6) as an indicator.  This ripeness ratio is measured as the percentage of red berries (class 4 and 5) on a collection date divided by the percentage of red berries on the final collection date. 


The Mullica Queen variety has a relatively low risk of overheating based on its albedo class distribution for the first four collection dates. On the fifth collection date of 9/9, the number of red berries significantly increases, indicating that the berries' overheating risk is now high. This pattern is observed with slight variations over all three Mullica Queen cranberry beds. The Stevens variety has a majority of green berries for a significant portion of the collection period. However, by 9/9 it begins to cross over into the category for a high risk of overheating. 

The Crimson Queen variety crosses into the high risk category by 8/25. (Green albedo values  in the late-season graphs are an artifact due to mis-classification of some leaf pixels as berries.)
The Haines variety crosses into the high risk category on 8/25 (in the sixth row) or on 8/31 (in the seventh row). 

From Figure \ref{fig:albedoOverTime}, we see that the Haines variety ripens the fastest. The next fastest ripening cranberry variety is Crimson Queen, followed by Mullica Queen. The Stevens variety is the slowest to ripen. These dates indicate a rough timeline indicating when cranberry farmers will need to monitor their crop more closely. These dates also serve as markers for when to focus more heavily on crop irrigation to mitigate overheating concerns.

\section{Conclusion}

Computer vision segmentation of photometrically calibrates images provides a real-time tool to assess crop health,  allow for expedient intervention, and compare crop varieties. We show the effectiveness of semantic segmentation on albedo characterization over time in cranberry crops. Weakly supervised semantic segmentation enables a convenient, effective localization  of cranberries  without the need for expensive pixel-wise labeling. We collect a time-series of weekly images from seven cranberry bogs over the course of two months using drones to 
create a labeled cranberry dataset with images throughout the entire cranberry growing cycle. This framework characterizes color development over time for  cranberries and provides  key insight into berry overheating risk and crop health. We create a timeline of albedo change that gives farmers the tools to make more informed irrigation choices to prevent crop rot and conserve resources. The resulting temporal signatures give important predictive power to the growers enabling choices among cranberry crop varieties and implications of those choices in best agriculture practices. 
The methodology can be automated for large scale crop evaluation to support new methods of high throughput phenotyping. 

\section{Acknowledgments}
This project was sponsored by the USDA NIFA AFRI, United States Award Number: 2019-67022-29922 and the SOCRATES NSF NRT \#2021628. We thank Michael King who supervised drone imagery collection for CRAID 2022.

\clearpage
\bibliographystyle{Frontiers-Harvard} 
\bibliography{egbib}

\end{document}